\documentclass[conference]{IEEEtran}
\IEEEoverridecommandlockouts
\usepackage{cite}
\usepackage{amsmath,amssymb,amsfonts}
\usepackage{algorithmic}
\usepackage{graphicx}
\usepackage{textcomp}
\usepackage{xcolor}
\usepackage{booktabs}
\usepackage{url}
\usepackage[hidelinks]{hyperref}

\def\BibTeX{{\rm B\kern-.05em{\sc i\kern-.025em b}\kern-.08em
    T\kern-.1667em\lower.7ex\hbox{E}\kern-.125emX}}

\begin{document}

\title{Flex-DDCM: Zero-Shot Task-Oriented Image Compression via Diffusion Trajectory Guidance}

\author{\IEEEauthorblockN{Yixin Gao, Runsen Feng, Xin Li\textsuperscript{*}, Zhibo Chen\textsuperscript{*}}
\IEEEauthorblockA{\textit{University of Science and Technology of China}\\
\{gaoyixin, fengruns\}@mail.ustc.edu.cn, \{xin.li, chenzhibo\}@ustc.edu.cn}
\thanks{\textsuperscript{*}Corresponding authors: Xin Li (xin.li@ustc.edu.cn) and Zhibo Chen (chenzhibo@ustc.edu.cn).}
}

\maketitle

\begin{abstract}
Recent zero-shot image compression methods exploit the powerful priors of pretrained diffusion models to 
produce perceptually pleasing reconstructions without compression-specific training. 
However, image compression serves not only human viewing but also machine analysis, while existing zero-shot codecs 
have limited flexibility in meeting the diverse reconstruction requirements of these applications.
In this paper, we propose Flex-DDCM, a framework for zero-shot task-oriented image compression via diffusion trajectory guidance. 
Building on Turbo-DDCM, we adapt the coded diffusion trajectory through task-guided noise selection, embedding the effects of task guidance into the compressed representation.
Guidance terms can be applied individually or combined, with timestep-dependent strengths 
controlling their influence throughout denoising. 
This design enables task adaptation without introducing additional decoding complexity. 
Experiments demonstrate flexible trade-offs among pixel fidelity, perceptual quality, and downstream task performance, with competitive performance against trained task-oriented codecs and faster decoding than Diff-ICMH.
\end{abstract}


\section{Introduction}
\label{sec:introduction}
Zero-shot image compression has recently shown promising perceptual reconstruction quality by reusing pretrained diffusion models, avoiding the need for compression-specific training~\cite{elata2025psc,theis2022lossy,ohayon2025ddcm,vonderfecht2025lossy}.
These methods differ in how they represent and encode image information.
For instance, DiffC~\cite{theis2022lossy}, subsequently made practical
by~\cite{vonderfecht2025lossy}, uses reverse-channel coding
(RCC)~\cite{theis2022algorithms} to progressively communicate noisy
samples of the image, followed by denoising to obtain a reconstruction.
DDCM~\cite{ohayon2025ddcm} instead steers reverse diffusion toward
the input image by selecting noise vectors from shared, reproducible
codebooks and encoding their indices. Turbo-DDCM~\cite{vaisman2026turboddcm}
efficiently combines multiple codebook vectors at each step and
encodes the selected combinations together with quantized coefficients,
reducing both the required number of denoising steps and the coding
overhead. However, image compression serves not only human viewing but also
machine analysis, giving rise to diverse requirements for pixel
fidelity, perceptual quality, and semantic information preservation.
Although zero-shot codecs offer perceptually pleasing reconstructions,
their flexibility in adapting to these diverse requirements remains
limited.

To meet these diverse requirements, researchers have explored how to balance different reconstruction objectives within a single compression model.
For example, MRIC~\cite{agustsson2023multi} and CRDR~\cite{CRDR} utilize conditional
generators to adjust the trade-off between pixel fidelity and perceptual
quality, with CRDR additionally supporting bitrate control.
Liu et al.~\cite{liu2025rate} extend such control to machine vision,
balancing pixel fidelity and downstream task performance through scalable
coding and reconstruction interpolation.
MPA~\cite{zhang2024allinone} supports reconstruction for pixel fidelity, perceptual quality, and machine vision through shared and task-specific feature paths.
For machine vision applications, some studies have explored image compression frameworks that support multiple downstream tasks~\cite{feng2022image,feng2023prompticm,guo2024unified,tang2025toward,yin2025unified,chen2023transtic,adapt-icmh,yu2026mtmoa,feng2025difficmh}. 
TransTIC~\cite{chen2023transtic} and Adapt-ICMH~\cite{adapt-icmh}
adapt pretrained codecs using lightweight prompts and adapters,
respectively, although each task requires separate adaptation.
MT-MoA~\cite{yu2026mtmoa} enables joint multi-task adaptation through
shared adapter pools and dynamic routing, supporting task-conditioned
reconstructions from a single bitstream.
Diff-ICMH~\cite{feng2025difficmh} combines generative priors with semantic
consistency training to support multiple downstream tasks from a single
codec and bitstream without task-specific adaptation, while maintaining
perceptual quality.
However, most of these methods require compression-specific training or fine-tuning.

\begin{figure}[t]
  \setlength{\abovecaptionskip}{3pt}
  \centering
  \includegraphics[width=\linewidth]{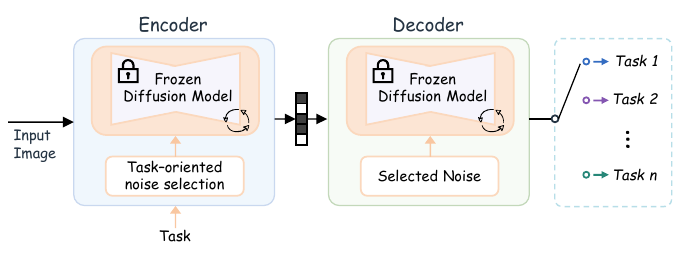}
  \caption{Overview of Flex-DDCM for task-oriented image compression.
  Task-guided noise selection during encoding enables adaptation to
  different task requirements, such as pixel fidelity and
  object detection performance. This adaptation introduces no additional decoding
  complexity.}
  \label{fig:intro}
\end{figure}

In this paper, we propose Flex-DDCM, a framework for zero-shot
task-oriented image compression via diffusion trajectory guidance.
Our key idea is to adapt the coded diffusion trajectory to the desired
reconstruction objective. In codebook-based diffusion compression, the
encoder determines this trajectory through a sequence of noise selections,
providing a means of embedding task preferences into the compressed
representation.
Building on Turbo-DDCM~\cite{vaisman2026turboddcm}, we develop a task-guided
selection criterion that combines the reconstruction residual with descent
directions derived from task objectives.
Different guidance terms can be applied individually or combined
to balance competing reconstruction objectives.
Motivated by the progression from large-scale structures to fine
details in diffusion generation~\cite{ho2020denoising}, we further
use timestep-dependent guidance strengths to emphasize task-specific
refinement at later denoising stages.
As illustrated in Fig.~\ref{fig:intro}, the resulting guidance is conveyed
through the transmitted noise selections, enabling task adaptation without
additional decoding complexity.

Our contributions are summarized as follows:
\begin{itemize}
  \item We propose Flex-DDCM, a framework for zero-shot task-oriented
  image compression that uses objective gradients to guide encoder-side
  noise selection, without additional decoding complexity.
  \item We develop a flexible guidance formulation that supports
  individual and combined reconstruction objectives, with
  timestep-dependent strengths controlling their influence throughout
  denoising.
  \item Experiments demonstrate flexible trade-offs among pixel fidelity,
  perceptual quality, and downstream task performance, with competitive
  performance against trained task-oriented codecs and faster decoding
  than Diff-ICMH~\cite{feng2025difficmh}.
\end{itemize}

\section{Method}
\label{sec:method}

\subsection{Task-guided Noise Selection}
\label{sec:flex-selection}

Flex-DDCM builds on Turbo-DDCM~\cite{vaisman2026turboddcm}.
Given an image $\mathbf{x}$ with VAE latent $\mathbf{z}_0=\mathcal{E}(\mathbf{x})$,
Turbo-DDCM selects $M$ atoms from a shared, reproducible Gaussian codebook
$\mathcal{C}_t\in\mathbb{R}^{d\times K}$ at each coded timestep, where $d$
is the latent dimension and $K$ is the codebook size. Selection uses the
largest absolute correlations with the residual
$\mathbf{r}_t=\mathbf{z}_0-\hat{\mathbf{z}}_{0|t}$, where
$\hat{\mathbf{z}}_{0|t}$ is the clean latent predicted from $\mathbf{z}_t$.
The selected atoms and their quantized coefficients form a sparse vector
$\mathbf{s}_t$ that determines the reverse update:
\begin{equation}
    \boldsymbol{\eta}_t=
    \frac{\mathcal{C}_t\mathbf{s}_t}
         {\operatorname{std}(\mathcal{C}_t\mathbf{s}_t)},
    \quad
    \mathbf{z}_{t-1}=\mu_t(\mathbf{z}_t)+\sigma_t\boldsymbol{\eta}_t,
    \label{eq:turboddcm_noise}
\end{equation}
where $\mu_t$ is the reverse mean and $\sigma_t$ is the noise scale.
Here, $t$ is the original diffusion timestep, and $t-1$ denotes the next
sampled timestep.
The selected indices are encoded as an unordered set with their
coefficients, allowing the decoder to reproduce the trajectory.

\begin{figure}[!t]
  \setlength{\abovecaptionskip}{3pt}
  \centering
  \includegraphics[width=\columnwidth]{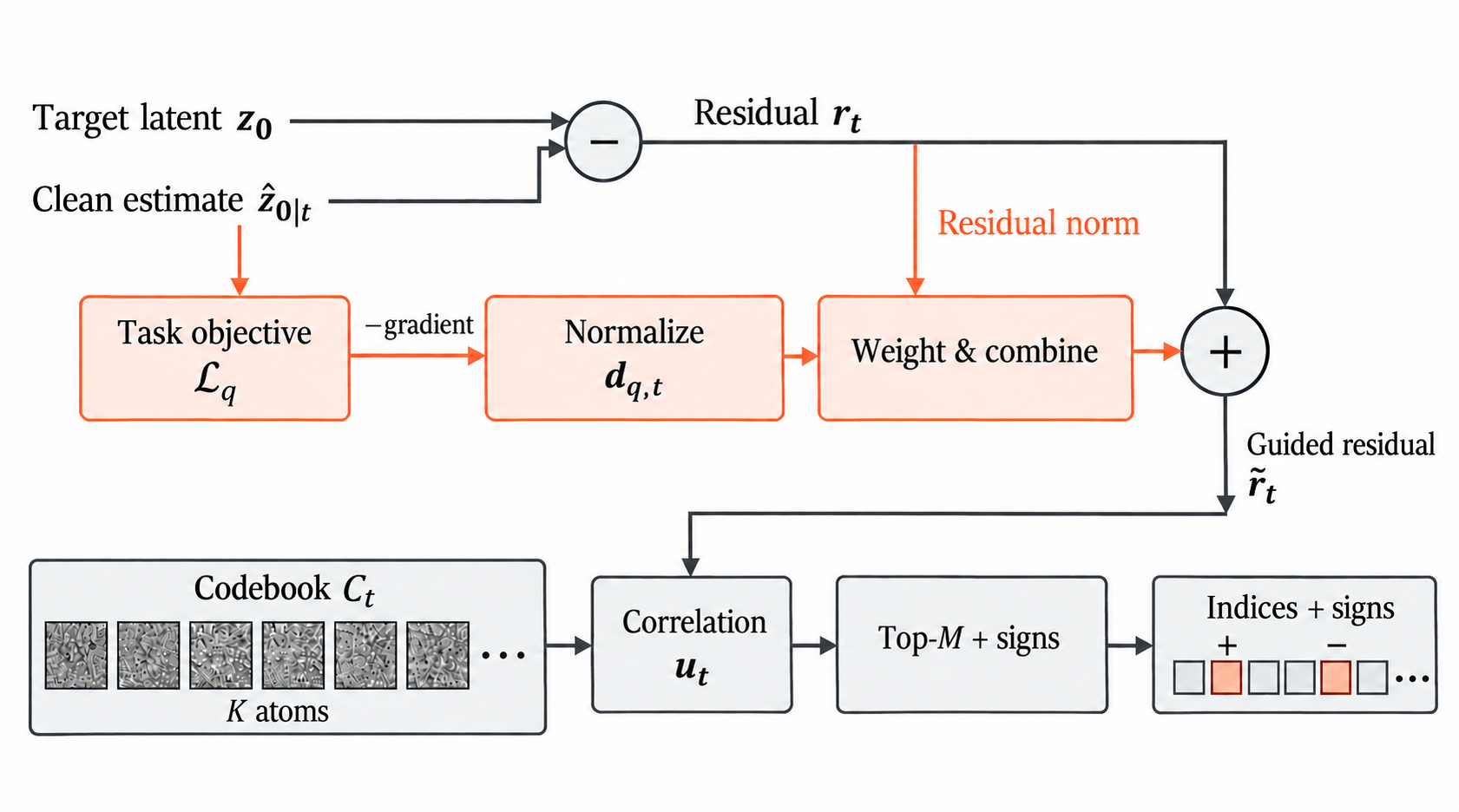}
  \vspace{-0.5cm}
  \caption{Task-guided noise selection in Flex-DDCM. Normalized task
  gradients are weighted and combined with the latent residual to select
  codebook atoms, whose indices and signs convey task guidance.}
  \label{fig:noise-selection}
\end{figure}
\begin{figure*}[!t]
  \centering
  \includegraphics[width=0.85\textwidth]{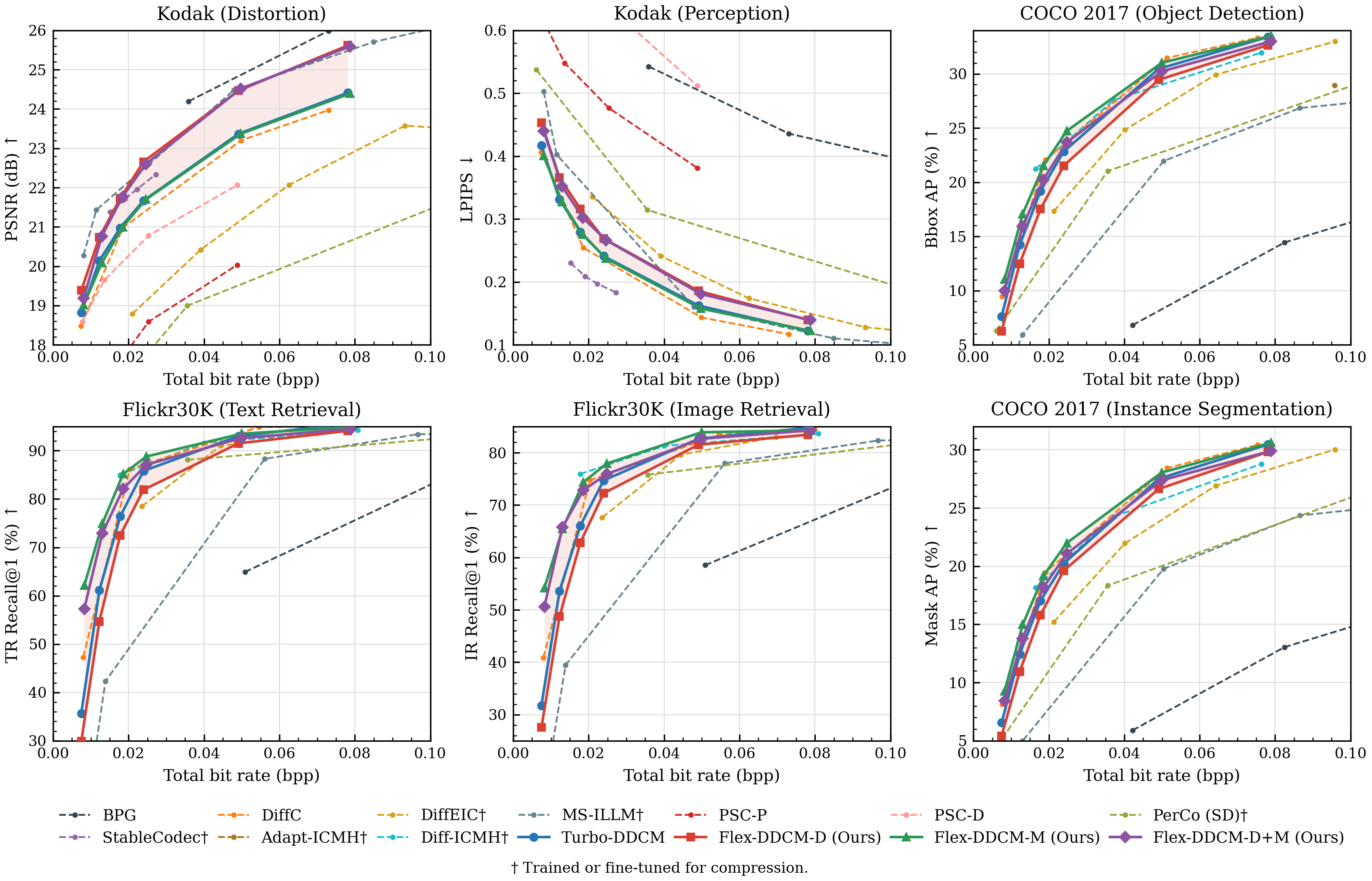}
  \caption{Performance comparison on Kodak, COCO 2017, and Flickr30K.
  Colored areas represent the adjustable range of Flex-DDCM.}
  \label{fig:main-results}
\end{figure*}

\begin{table*}[!t]
  \centering
  \caption{Average runtime on $512\times512$ Kodak images using an NVIDIA RTX~4090. Enc. and Dec. denote encoding and decoding time. \\ $^\dagger$ Trained or fine-tuned for compression.}
  \label{tab:runtime}
  \small
  \setlength{\tabcolsep}{3pt}
  \begin{tabular*}{0.9\textwidth}{@{\extracolsep{\fill}}l*{8}{c}@{}}
    \toprule
     & DiffC & PerCo (SD)$^\dagger$ & DiffEIC$^\dagger$ & Diff-ICMH$^\dagger$ & Turbo-DDCM & Flex-DDCM-D & Flex-DDCM-M & Flex-DDCM-D+M \\
    \midrule
    \textbf{Enc. (s)} & 3.560 & 0.365 & 0.129 & 0.160 & 1.068 & 1.939 & 1.631 & 2.537 \\
    \textbf{Dec. (s)} & 3.210 & 1.170 & 3.291 & 6.466 & 0.986 & 0.994 & 1.032 & 1.033 \\
    \bottomrule
  \end{tabular*}
\end{table*}

As shown in Fig.~\ref{fig:noise-selection}, Flex-DDCM adjusts the residual
used for atom selection with local descent directions of differentiable task objectives $\mathcal{L}_q(\mathbf{z};\mathbf{x})$.
At the current clean prediction, we compute
\begin{equation}
    \begin{aligned}
    \mathbf{g}_{q,t}&=\left.\nabla_{\mathbf{z}}
        \mathcal{L}_q(\mathbf{z};\mathbf{x})
        \right|_{\mathbf{z}=\hat{\mathbf{z}}_{0|t}},\\
    \mathbf{d}_{q,t}&=-\frac{\mathbf{g}_{q,t}}
        {\|\mathbf{g}_{q,t}\|_2+\epsilon},
    \end{aligned}
    \label{eq:flex-local-gradient}
\end{equation}
and combine the normalized directions with the latent residual:
\begin{equation}
    \tilde{\mathbf{r}}_t=\mathbf{r}_t+
    (\|\mathbf{r}_t\|_2+\epsilon)
    \sum_{q\in\mathcal{Q}}a_q\gamma_q w_{q,t}\mathbf{d}_{q,t}.
    \label{eq:flex-composed-objective}
\end{equation}
Here, $\mathcal{Q}$ contains the active objectives, $a_q\geq0$ weights their
relative contributions, $\gamma_q\geq0$ sets each guidance strength,
$w_{q,t}$ specifies its temporal profile, and $\epsilon$ ensures numerical
stability. Residual-norm scaling makes the guidance strength relative to
the reconstruction residual. A single objective uses
$\mathcal{Q}=\{q\}$ and $a_q=1$.

We select the $M$ atoms most correlated with the guided residual in
absolute value and assign signed unit coefficients:
\begin{equation}
    \begin{aligned}
    \mathbf{u}_t&=\mathcal{C}_t^\top\tilde{\mathbf{r}}_t,
    &\mathcal{I}_t&=\operatorname{TopM}(|\mathbf{u}_t|),\\
    (\mathbf{s}_t)_k&=
    \begin{cases}
        \operatorname{sign}(u_{t,k}),&k\in\mathcal{I}_t,\\
        0,&k\notin\mathcal{I}_t,
    \end{cases}
    \end{aligned}
    \label{eq:flex-guided-selection}
\end{equation}
where $\operatorname{TopM}$ returns the indices of the $M$ largest entries.
The noise and reverse update follow~\eqref{eq:turboddcm_noise}.
Task guidance is thus conveyed through the transmitted indices and signs;
the decoder evaluates neither task objectives nor their gradients,
introducing no additional decoding complexity.

\subsection{Guidance Scheduling}
\label{sec:flex-scales}
Motivated by the progression from coarse structures to fine details in
diffusion generation~\cite{ho2020denoising}, we vary guidance strength
across timesteps. Each objective uses one of three profiles:
\begin{equation}
    h_q(t)=
    \begin{cases}
        1,&\text{constant},\\
        1/\max(t,1),&\text{inverse timestep},\\
        \sqrt{\bar\alpha_t/(1-\bar\alpha_t)},&\text{square-root SNR},
    \end{cases}
    \label{eq:flex-scale-profiles}
\end{equation}
where $\bar\alpha_t$ is the cumulative signal coefficient and SNR denotes
signal-to-noise ratio. The latter two profiles emphasize later denoising
steps. We normalize over the timesteps $\mathcal{T}_{\mathrm{bit}}$ at
which codebook indices are transmitted:
\begin{equation}
    w_{q,t}=\frac{h_q(t)}{
        \frac{1}{|\mathcal{T}_{\mathrm{bit}}|}
        \sum_{s\in\mathcal{T}_{\mathrm{bit}}}h_q(s)}.
    \label{eq:flex-scale-normalization}
\end{equation}
Each profile then has unit mean, so $\gamma_q$ controls the average guidance
strength. Section~\ref{sec:flex-ablation} compares the schedules.

\subsection{Task Objectives}
\label{sec:flex-objectives}

For pixel fidelity (D), we use mean squared reconstruction error:
\begin{equation}
    \mathcal{L}_{D}(\mathbf{z};\mathbf{x})=
    \frac{1}{3HW}\|\mathcal{D}(\mathbf{z})-\mathbf{x}\|_2^2,
    \label{eq:flex-distortion-objective}
\end{equation}
where $\mathcal{D}$ is the VAE decoder and $H,W$ are the image dimensions.
For perceptual quality (P), we use only the latent residual for selection,
without task-gradient guidance, as in Turbo-DDCM~\cite{vaisman2026turboddcm}.

For machine vision (M), following Diff-ICMH~\cite{feng2025difficmh}, we
use semantic consistency to preserve features shared across downstream
tasks. Let $\phi(\mathbf{z})$ be the middle-block feature map of the frozen
diffusion UNet, evaluated on the clean latent at timestep $0$ with
null-text conditioning. We define the semantic guidance objective as the average cosine distance:
\begin{equation}
    \mathcal{L}_{M}(\mathbf{z};\mathbf{x})=
    1-\frac{1}{P}\sum_{p=1}^{P}
    \operatorname{cos}\bigl(\phi_p(\mathbf{z}),\phi_p(\mathbf{z}_0)\bigr),
    \label{eq:flex-machine-objective}
\end{equation}
where $\phi_p$ denotes the feature vector at spatial location $p$, $P$ is
the number of locations, and $\operatorname{cos}$ is cosine similarity.
Reference features $\phi(\mathbf{z}_0)$ are computed once and cached.
The D+M configuration jointly applies distortion and semantic guidance
using $\mathcal{Q}=\{D,M\}$ in~\eqref{eq:flex-composed-objective}, with
separate strengths and schedules.

\begin{figure*}[!t]
  \setlength{\abovecaptionskip}{3pt}
  \centering
  \includegraphics[width=0.72\textwidth]{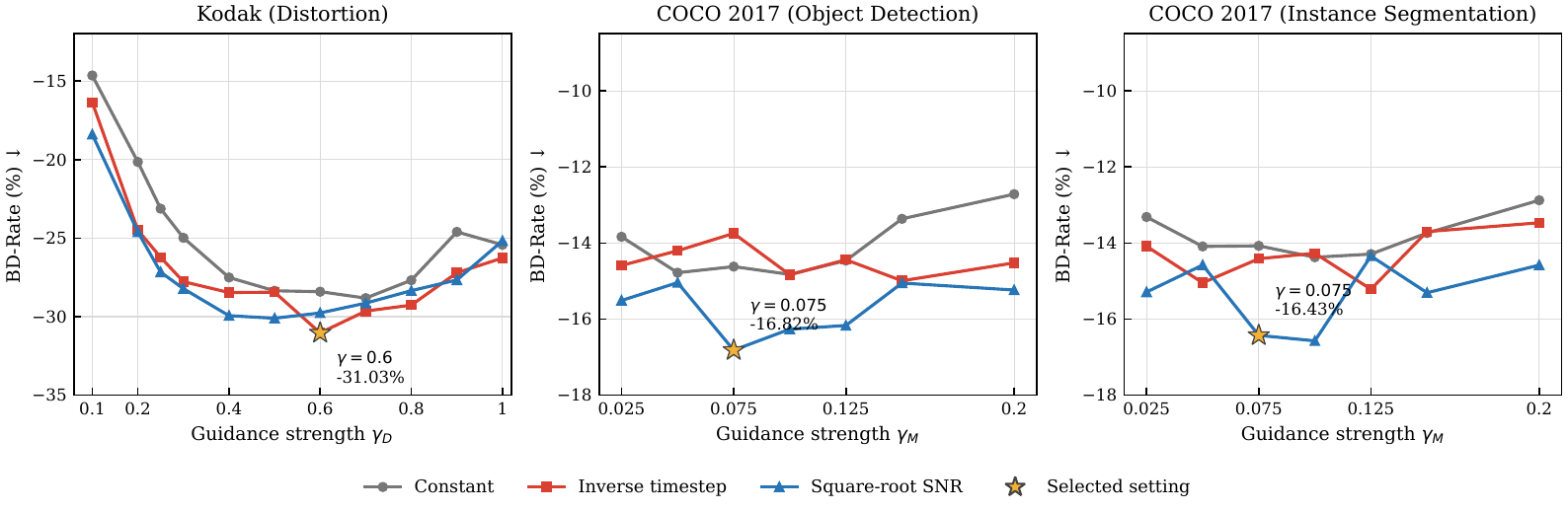}
  \caption{Effect of guidance strength and temporal scheduling on Kodak distortion and COCO 2017 object detection and instance segmentation.}
  \label{fig:guidance-schedule}
\end{figure*}

\section{Experiments}
\label{sec:evaluation}

\subsection{Experimental Settings}
\paragraph{Evaluation Datasets and Metrics}
We evaluate our approach in terms of reconstruction quality and downstream
machine vision performance on three public datasets.
On Kodak~\cite{kodak}, we assess signal fidelity using PSNR and
perceptual quality using LPIPS~\cite{lpips}. 
We evaluate object detection and instance segmentation on COCO 2017~\cite{coco}
using Faster R-CNN~\cite{ren2015faster} and Mask R-CNN~\cite{he2017mask},
respectively, both with R50-FPN backbones in Detectron2~\cite{wu2019detectron2}.
We report bounding-box AP and mask AP for the two tasks.
On Flickr30K, we use BEiT-3~\cite{wang2023beit3} for multimodal retrieval and report Recall@1
in both directions: image-to-text (text retrieval) and text-to-image
(image retrieval).
Following prior work~\cite{ohayon2025ddcm,vaisman2026turboddcm}, we
center-crop Kodak images to $512\times512$ without resizing.
COCO 2017 and Flickr30K images are directly resized to $512\times512$
using bicubic interpolation.

\paragraph{Baseline Methods}
We compare our method with zero-shot diffusion-based codecs, including
DiffC~\cite{vonderfecht2025lossy},
Turbo-DDCM~\cite{vaisman2026turboddcm}, and
PSC~\cite{elata2025psc} with distortion- and perception-oriented
reconstruction (PSC-D and PSC-P). These methods reuse pretrained
diffusion models without compression-specific training. The P configuration
is represented by Turbo-DDCM in the comparisons. We also compare with the
traditional codec BPG~\cite{bellard_bpg} and codecs trained or fine-tuned for
image compression, including PerCo~\cite{perco}, MS-ILLM~\cite{msillm},
DiffEIC~\cite{li2024towards}, and StableCodec~\cite{stablecodec}.
For machine vision, we further include the task-oriented codecs
Adapt-ICMH~\cite{adapt-icmh} and Diff-ICMH~\cite{feng2025difficmh}.

\paragraph{Implementation Details}
We use pretrained Stable Diffusion 2.1-base~\cite{rombach2022high}
with $T=30$ sampling steps, $K=16{,}384$, $C=1$ bit per coefficient ($\pm1$), and
$M$ ranging from $45$ to $100$.
We adopt Turbo-DDCM's~\cite{vaisman2026turboddcm} timestep schedule and
bitrate-dependent $\mathcal{T}_{\mathrm{bit}}$, followed by uncoded
deterministic DDIM steps ($\eta=0$).
Flex-DDCM-D uses MSE guidance ($\gamma_D=0.6$, inverse-timestep weighting),
while Flex-DDCM-M uses semantic guidance ($\gamma_M=0.075$, square-root
SNR weighting) and RAM++~\cite{huang2025ramplusplus} tag conditioning.
Flex-DDCM-D+M combines both terms with $a_D=a_M=1$ and the same
strengths, schedules, and tag conditioning.
Following Diff-ICMH~\cite{feng2025difficmh}, we losslessly encode and
transmit tags as 13-bit indices into a shared dictionary; the decoder
recovers comma-separated text for conditioning. Flex-DDCM bpp uses the
actual image and tag file sizes in bits, including headers and padding,
divided by the image pixel count.

\subsection{Main Results}

\paragraph{Quantitative Results}
Fig.~\ref{fig:main-results} shows how task-oriented noise selection
controls the trade-off between reconstruction fidelity and downstream
task performance. Relative to Turbo-DDCM, distortion guidance
consistently improves PSNR. Semantic guidance improves downstream
performance, particularly at low bitrates, while maintaining comparable
LPIPS. Combining the two objectives allows Ours-D+M to retain most of
the PSNR gains of Ours-D while improving its downstream performance. 
Ours-M is also competitive with trained task-oriented codecs.
Compared with Diff-ICMH~\cite{feng2025difficmh}, it achieves comparable
text and image retrieval performance on Flickr30K and higher detection
and segmentation AP on COCO 2017 at moderate and high bitrates,
without requiring compression-specific training.

\paragraph{Runtime}
Table~\ref{tab:runtime} shows that encoder-only task guidance increases
encoding time while retaining Turbo-DDCM's decoding procedure.
Despite this encoding overhead, Flex-DDCM supports different reconstruction
objectives while retaining efficient decoding. For machine vision,
Flex-DDCM-M decodes substantially faster than Diff-ICMH~\cite{feng2025difficmh}
while maintaining competitive downstream performance.

\subsection{Ablation Studies}
\label{sec:flex-ablation}

\begin{table}[!t]
\centering
\caption{Ablation of tag conditioning and classifier-free guidance (CFG) on COCO 2017. BD-rate is relative to Flex-DDCM-M (lower is better).}
\label{tab:tags-cfg-ablation}
\begin{tabular*}{0.92\columnwidth}{@{\extracolsep{\fill}}lrr@{}}
\toprule
\textbf{Variant} & \textbf{Detection (\%)} & \textbf{Segmentation (\%)} \\
\midrule
w/o tags & +8.06 & +7.32 \\
w/ CFG & +22.37 & +23.04 \\
\midrule
\textbf{Flex-DDCM-M (Ours)} & \textbf{0.00} & \textbf{0.00} \\
\bottomrule
\end{tabular*}
\end{table}

\paragraph{Guidance Strength and Schedule}
As shown in Fig.~\ref{fig:guidance-schedule}, inverse-timestep guidance
achieves the largest bitrate savings for distortion, while square-root
SNR guidance performs best for detection and segmentation.
Both improve over the best constant-guidance setting.
The optimal strengths are $\gamma_D=0.6$ for distortion and
$\gamma_M=0.075$ and $0.1$ for detection and segmentation, respectively.

\paragraph{Tag Conditioning and Classifier-Free Guidance}
As shown in Table~\ref{tab:tags-cfg-ablation}, removing tags increases
BD-rate for both object detection and instance segmentation.
Following Diff-ICMH~\cite{feng2025difficmh}, we set the CFG scale to $5.0$
for the CFG variant.
Among the evaluated settings, RAM++ tag conditioning without CFG
achieves the lowest BD-rate on both tasks.

\section{Conclusion}
\label{sec:conclusion}
We presented Flex-DDCM to adapt zero-shot diffusion compression to diverse reconstruction requirements. 
Task-objective gradients guide encoder-side noise selection, conveying task preferences 
through the compressed representation without introducing additional decoding complexity. 
Guidance terms can be used individually or combined, with timestep-dependent strengths controlling their influence during denoising. 
Experiments demonstrate flexible trade-offs among pixel fidelity, perceptual quality, and downstream task performance. 
Flex-DDCM also achieves competitive downstream performance against trained task-oriented codecs, with faster decoding than Diff-ICMH.

\bibliographystyle{IEEEtran}
\bibliography{main}

\end{document}